# Automated Deep Learning Estimation of Anthropometric Measurements for Preparticipation Cardiovascular Screening

Lucas R. Mareque, Ricardo L. Armentano, and Leandro J. Cymberknop

Universidad Tecnológica Nacional, Facultad Regional Buenos Aires, Grupo de Investigación y Desarrollo en Bioingeniería (GIBIO), Ciudad Autónoma de Buenos Aires, Argentina
`Lmareque@frba.utn.edu.ar`
`Gibio@frba.utn.edu.ar`

**Abstract.** Preparticipation cardiovascular examination (PPCE) aims to prevent sudden cardiac death (SCD) by identifying athletes with structural or electrical cardiac abnormalities. Anthropometric measurements, such as waist circumference, and also limb lengths, and torso proportions to detect Marfan syndrome, can indicate elevated cardiovascular risk. Traditional manual methods are labor-intensive, operator-dependent, and challenging to scale. We present a fully automated deep-learning approach to estimate five key anthropometric measurements from 2D synthetic human body images. Using a dataset of 100,000 images derived from 3D body meshes, we trained and evaluated VGG19, ResNet50, and DenseNet121 with fully connected layers for regression. All models achieved sub-centimeter accuracy, with ResNet50 performing best, achieving a mean MAE of 0.668 cm across all measurements. Our results demonstrate that deep learning can deliver accurate anthropometric data at scale, offering a practical tool to complement athlete screening protocols. Future work will validate the models on real-world images to extend applicability.

## 1 Introduction

The preparticipation cardiovascular examination (PPCE) is recommended by the world's leading medical societies and is fundamentally aimed at preventing sudden cardiac death (SCD), with the goal of reducing the incidence of this event. [1-3]. Essentially, the aim is to identify athletes, whether recreational or competitive, who may have an unsuspected structural or electrical cardiac condition. It has been shown that exercise can act as a trigger, which, in the presence of an undetected condition, may lead to SCD [4]. Although the incidence of sudden cardiac death varies according to race and type of sporting activity [5], it is estimated to range from 0.5 to 2 individuals per 100,000 people per year. Regarding the causes, hereditary conditions such as hypertrophic cardiomyopathy (predominant) or arrhythmogenic right ventricular dysplasia are considered, although around 5% of cases remain unexplained. Regarding age, it peaks around the fourth decade of life (with a risk of SCD of 4 to 5 per 100,000 people per year), during which a significant number of cardiomyopathies and coronary artery diseases are reported, with coronary artery disease being the most prevalent [6].

Among the aspects to consider, anthropometry plays a fundamental role in identifying morphological alterations that can impact the risk of SCD, including abnormal body fat distribution, disproportionate chest or torso dimensions, extreme limb lengths, or other structural deviations, such as those observed in conditions like Marfan syndrome, which may indicate underlying cardiovascular or musculoskeletal abnormalities [7].

Traditionally, anthropometric data collection relies on manual methods involving instruments like anthropometers, and tape measures, often requiring standardized postures and precise calibration. However, these approaches are labor-intensive, time-consuming, and their accuracy depends heavily on the skill of the operator and correct placement of measuring tools. This introduces variability and limits scalability, especially when dealing with large populations [8].

To overcome these challenges, and considering that current strategies for sudden cardiac death prevention focus on early detection (prior to engaging in physical activity), this study introduces a fully automated and precise approach for estimating human anthropometric measurements from two-dimensional images. This work contributes to the field by presenting a deep-learning-based pipeline capable of automatically estimating key anthropometric measurements from 2D synthetic images, paving the way for scalable and accurate human body analysis applications.

## 2  Materials and Methods

### 2.1  Dataset

The model was trained on a large-scale dataset of synthetically generated human body data, comprising 50.000 female and 50.000 male 2D projections derived from 3D body meshes in the standard T-pose (see Fig. 1). This dataset, obtained from Skeletex Research[9], was created using the Skinned Multi-Person Linear (SMPL) parametric model [10]. It contains a wide variety of body shapes representations with grayscale images, and a set of 16 annotated anthropometric measurements.

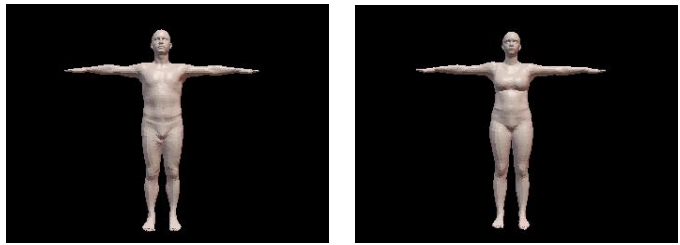

**Fig. 1.** On the left, a sample from the synthetic male population; on the right, a sample from the synthetic female population

The grayscale images were produced by rendering the 3D meshes from a frontal viewpoint, while ensuring consistent alignment in T-pose with the Y-axis representing the vertical direction and the Z-axis facing the camera. Table 1 summarizes the definitions of the annotated measurements as defined by the authors relevant to this work.

Table 1. Definition of each body measurement type according to the authors.[9]

| Measurement | Definition |
|---|---|
| Shoulder-to-wrist | distance between the shoulder and the wrist joint (sleeve length) |
| Torso length | distance between the neck and the pelvis joint |
| Waist circumference | circumference taken at the Y-axis level of the minimal intersection of a model and the mesh signature within the waist region – around the natural waist line (mid-spine joint); the region is scaled relative to the model stature |
| Pelvis circumference | circumference taken at the Y-axis level of the maximal intersection of a model and the mesh signature within the pelvis region, constrained by the pelvis joint and hip joint |
| Leg length | distance between the pelvis and ankle joint |

## 2.2 Anthropometric Dimension Calculation Based on transfer learning

We chose several well-known deep learning models pre-trained on ImageNet[11] for comparison to serve as backbones for feature extraction, combined with a fully connected neural network to predict each measurement (see Fig. 1). Specifically, VGG19[12], ResNet50[13], and DenseNet121[14] were used due to their demonstrated effectiveness across medical domains [15]:

- VGG19: with its 19 layers (comprising 16 convolutional layers and 3 fully connected layers) utilizes small 3x3 convolution filters to capture detailed features effectively. Its use in medical imaging, often via transfer learning, has demonstrated notable success [16].

- ResNet50: revolutionized image classification by introducing residual learning to address the vanishing gradient issue in deep networks. The use of skip connections allows input signals to bypass certain layers and merge later, preserving information and stabilizing training. ResNet50, in particular, has been widely applied in medical imaging through transfer learning [17].

- DenseNet121: features a unique densely connected architecture where each layer receives input from all previous layers, enhancing information flow and gradient propagation. This promotes better feature reuse, reducing the number of parameters while maintaining high performance.

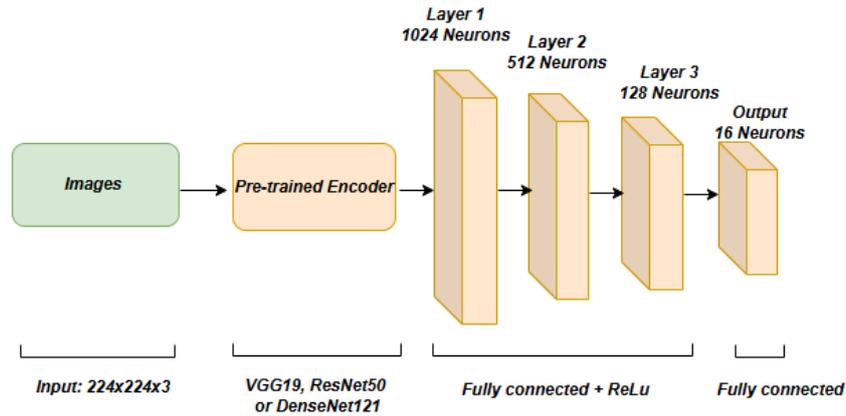

**Fig. 1.** Architecture of the transfer learning models used

After obtaining anthropometric measurements, several metrics can be calculated to evaluate cardiovascular risk and detect potential connective tissue disorders.

Waist Circumference is a measure of central obesity and a stronger predictor of metabolic syndrome and cardiovascular events than BMI alone. Recommended cut-off points from World Health Organization [18] are:

- Men: Increased risk ≥ 94 cm; high risk ≥ 102 cm
- Women: Increased risk ≥ 80 cm; high risk ≥ 88 cm

Individuals exceeding these thresholds should be evaluated for blood glucose, lipid profile, and blood pressure before engaging in vigorous physical activity.

Waist-to-Hip Ratio (WHR) is a metric that indicates where body fat is distributed, with a higher WHR suggesting a greater proportion of fat around the abdomen (android pattern), which is associated with increased cardiovascular risk. For men, a WHR above 0.90 is considered to indicate increased risk, while for women, the threshold is a WHR above 0.85 [18].

Additionally, anthropometric data can be used to detect Marfanoid habitus, indicative of potential connective tissue disorders such as Marfan syndrome. Key measurements include:

- Length of torso
- Length of legs

- Length of arms

These measurements are important because arms and legs may be disproportionately long when compared with the torso [19].

### 2.3 Training details

The dataset consisted of 100.000 RGB images (50.000 male and 50.000 female) and were split into 70.000 images for training, 15.000 for validation, and 15.000 for testing. The test set was further separated into male and female subsets to enable gender-specific evaluation. The training set was created by combining male and female samples in equal proportions, and all datasets were randomly shuffled with a fixed seed to ensure reproducibility.

Images were preprocessed according to the requirements of the selected backbone architecture. Each image was decoded in RGB format, resized to 224×224 pixels, and normalized using the corresponding ImageNet-specific preprocessing function.

For each backbone architecture, the convolutional base was frozen during training, followed by a Flatten layer and three fully connected layers with 1024, 512, and 128 units, respectively, each followed by Batch Normalization. The output layer consisted of 16 neurons for regression.

Training was performed using the Adam optimizer with a learning rate of $1x10^{-4}$ and mean absolute error (MAE) as the loss function. Early stopping with a patience of 10 epochs was applied to prevent overfitting, restoring the best-performing weights based on validation loss. All models were trained for a maximum of 100 epochs with a batch size of 350 images. Although the models were trained to predict all 16 annotated measurements for the original dataset, only five were considered relevant and are reported in this work.

All experiments were implemented in TensorFlow/Keras and executed on an NVIDIA A100 GPU using Google Colab® (Google LLC, Californa, USA).

## 3 Results

The evaluation results are presented in Table 2, which summarizes the Mean Absolute Error (MAE, in centimeters) for each relevant measurement to this work on the male, female, and combined test datasets. For ResNet50, the "Total MAE" column for the combined dataset highlights measurements of arm length and torso length that perform better compared to Conv-BoDiEs[9], which was trained and evaluated on the same dataset used in this work.

Overall, ResNet50 achieved a Mean MAE of 0.668 cm across all measurements and subjects, with slightly better performance on the male dataset (0.665 cm) than on the female dataset (0.666 cm). As shown in Table 3, ResNet50 slightly outperformed VGG19 and DenseNet121 in overall accuracy.

**Table 2.** Mean Absolute Error (MAE, cm) for each model

| Measurement | VGG19 | | | ResNet50 | | | DenseNet121 | | |
|---|---|---|---|---|---|---|---|---|---|
| | Male MAE | Female MAE | Total MAE | Male MAE | Female MAE | Total MAE | Male MAE | Female MAE | Total MAE |
| Waist circumference | 1.283 | 1.416 | 1.359 | 1.178 | 1.283 | 1.234 | 1.212 | 1.360 | 1.288 |
| Pelvis circumference | 1.192 | 1.123 | 1.157 | 0.965 | 0.933 | 0.953 | 1.041 | 1.067 | 1.044 |
| Arm length | 0.361 | 0.368 | 0.361 | 0.366 | 0.393 | **0.379** | 0.354 | 0.394 | 0.376 |
| Leg length | 0.436 | 0.416 | 0.430 | 0.382 | 0.354 | 0.375 | 0.421 | 0.381 | 0.401 |
| Torso length | 0.501 | 0.404 | 0.451 | 0.434 | 0.367 | **0.398** | 0.452 | 0.426 | 0.449 |

**Table 3.** Mean MAE (cm) across all measurements

| Model | Male Mean MAE | Female Mean MAE | Total Mean MAE |
|---|---|---|---|
| VGG19 | 0.755 | 0.745 | 0.752 |
| ResNet50 | 0.665 | 0.666 | 0.668 |
| DenseNet121 | 0.696 | 0.726 | 0.710 |

## 4    Discussion

Preparticipation cardiovascular evaluation aims to identify athletes with conditions that may predispose them to sudden cardiac death (SCD), distinguishing between physiological adaptations to training and pathological findings such as cardiomyopathies [6],[20]. Anthropometric assessment is particularly relevant, as certain body proportions, such as waist circumference, arm length, or marfanoid habitus indicators, can signal underlying cardiovascular or connective tissue disorders [1],[3].

In this study, we propose an automated deep learning approach to estimate five anthropometric measurements from synthetically generated 2D human body images. Using the waist and pelvis circumferences, the Waist-to-Hip Ratio (WHR) can also be calculated. All three evaluated architectures (VGG19, DenseNet121, ResNet50) achieved sub-centimeter accuracy, with ResNet50 performing best (total MAE = 0.668 cm). Notably, these models surpassed Conv-BoDiEs[9] in some measurements, even though Conv-BoDiEs[9] was trained from scratch, whereas our approach leveraged transfer learning. This level of precision suggests that deep learning models can provide reliable, reproducible anthropometric data without the time, equipment, and operator dependency of traditional manual methods.

The use of a synthetic dataset allowed precise ground-truth labels and wide variation in body shapes, but it also limits direct generalization to real-world scenarios where image conditions vary. Future work should validate and fine-tune the models on real annotated photographs, incorporating techniques to handle occlusions, clothing, and non-standard poses.

Overall, these results demonstrate the feasibility of automated anthropometric estimation from 2D images, offering a scalable tool to complement current athlete screening protocols and potentially enhance early detection of individuals at elevated cardiovascular risk.